# Improve Document Embedding for Text Categorization Through Deep Siamese Neural Network

**Erfaneh Gharavi**

Data Science Institute, University of Virginia, Charlottesville, Virginia, USA

eg8qe@virginia.edu

**Hadi Veisi** *

Data and Signal Processing Lab, Faculty of New Sciences and Technologies, University of Tehran, Tehran, Iran

h.veisi@ut.ac.ir

+989126214303

## Abstract

Due to the increasing amount of data on the internet, finding a highly-informative, low-dimensional representation for text is one of the main challenges for efficient natural language processing tasks including text classification. This representation should capture the semantic information of the text while retaining their relevance level for document classification. This approach map the documents with similar topics to a similar space in vector space representation. To obtain representation for large text, we propose the utilization of deep Siamese neural networks. To embed document relevance in topics in the distributed representation, we use a Siamese neural network to jointly learn document representations. Our Siamese network consists of two sub-network of multi-layer perceptron. We examine our representation for the text categorization task on BBC news dataset. The results show that the proposed representations outperform the conventional and state-of-the-art representations in the text classification task on this dataset.

## 1 Introduction

With the rapid growth of online information, particularly in text format, text classification has become a significant technique for managing this type of data [1]. Text categorization, or text classification is the process of assigning a label to a document. This task has been widely studied and addressed in many real applications [2]–[5] over the last few decades.

Unstructured text must be converted into a structured format when using mathematical modeling as part of the classification process. Semantic representations [6]–[8] of large documents for document classification and retrieval tasks have captured researchers' attention. Currently, several of the most popular representation methods for documents are bag-of-words models. These methods require high-dimensional representations, and they are computationally expensive because of the high cost of float/integer computation. A low dimensional representation could decrease the computational complexity of training document classification models.

One new approach for creating semantic representations involves training deep neural networks to create vectors representing documents. Previous work has shown the advantages of using deep neural network approaches rather than bag-of-word approaches to create document representations [9]–[11].

To facilitate document categorization, documents must have comparable representations that embed their relevancy within the document topics. Deep Siamese neural networks can be used to jointly learn similar document representations for documents with the same topic. The first application of Siamese neural networks was for testing the similarity between two signatures [12]. The architecture of Siamese neural networks makes them applicable when there is similarity between two objects. Siamese neural networks are composed of two sub-networks. If the type of input for each sub-network is the same, the sub-networks could have the similar structure. The outputs of these sub-networks are concatenated and input into the joint layer. This architecture makes it possible for the network to extract features from two similar inputs and assign a similarity score using the extracted feature vectors.

*Corresponding author

This paper is inspired by our previous work [13], and we use a similar general approach to build representations for the text categorization task without the utilization of a rhetorical structure tree. The goal of this paper is to create a method for training document representations that is able to embed relevancy information within a document topic. This can be done by employing a Siamese neural network with multi-layer perceptron networks as sub-networks to find deep informative low-dimensional representations for documents. This approach helps us to generate thousands of samples from few documents for training deep neural networks.

The rest of this paper is organized as follows: the Background and related work section describes conventional document representations in addition to recent papers on different text representation methods. The proposed method section discusses the suggested approach for text representation improvement. The experimental section explains our settings, dataset, and results. This is followed by the conclusion.

## 2 Background & Related Work

In this section, we explain the conventional methods followed by state-of-the-art document representations:

- **TFIDF**: Term frequency-inverse document frequency is one of the long-lasting methods for text representation. Though newer methods of document representation have been heavily explored in the past decade, this method remains popular specifically for large documents where word order does not matter [14].
- **LSA**: Latent semantic analysis is an extension of TFIDF where singular-value decomposition (SVD) is applied to the term-document matrix with the intuition that these new columns represent document topics [15]. LSA was applied to leave a specific number of columns equivalent to the document topics in the output matrix.
- **LDA**: Latent Dirichlet allocation builds a probabilistic distribution of topics in documents. Methods such as variational inference and Gibbs' Sampling are used to generate distributions assuming a Dirichlet prior for distribution of words and topics within the text [16].
- **Word Averaging:** For each document, we calculated this representation by averaging the word vectors [17]–[19] in the document as $wv$ in (1). Here, n is the number of words in the document and $w_j$ is the word vector of the $j^{th}$ word.

$$wv = \sum_j^n \frac{w_j}{n} \quad (1)$$

We use pre-trained Glove [20][2] vectors to represent words. Stop words are removed because their inclusion makes the average vectors indistinguishable from one-another.

In addition to these conventional methods, several state-of-the-art methods have recently been proposed: Salakhutdinov & Hinton [21] trained a deep graphical model of the word-count. This graphical model performs as semantic hashing since the deepest layer is forced to use small binary numbers. This semantic representation mapped similar documents to similar memory addresses. Wang et al [22] used Tags and Topic Modeling (SHTTM) to generate a semantic hashing representation. Livermore et al [23] , use a representation learning based on a combination of Topic Modeling and Citation Networks to retrieve the latent relevance structure of the documents. Exploiting their proposed Markov decision process (MDP) formulation of search, they learn process model of information retrieval strategies over the representation, through a reinforcement learning approach.

Authors in [24], they jointly learn the representation of the queries and documents in the latent vector space by forcing them to learn a high score for relevant query and documents and a low score for irrelevant ones. Lioma et al. [25] consider rhetorical relation in information retrieval and achieve significant improvement (> %10 in mean average precision) over a state-of-the-art baseline. Mueller, [26] use a Siamese adaptation of the Long Short-Term Memory (LSTM) network for labelled data comprised of pairs of variable-length sequences and evaluate semantic similarity between sentences. Ji and Smith [27] use RNN and an attention mechanism to compute text representations that focus on the nucleus of the text. Yin et al

[2] https://nlp.stanford.edu/projects/glove/

[28] present a Siamese convolutional neural network with attention mechanism to model sentence pairs similarity. Wang et al [30] utilize a neural network to represent texts as distributed vectors and evaluate the representation by clustering the labelled data representations. Bowman et al., [31] introduce an RNN-based auto-encoder to represent latent distributions of entire sentences. Sinoara et al., [32] construct the document representation by bring together word sense disambiguation and the semantic richness of word- and word-sense embedded vectors. By employing denoising autoencoder (DAE) and restricted Boltzmann machine (RBM), Aziguli et al., [33] propose method, named denoising deep neural network (DDNN), to achieve significant improvement with better performance of anti-noise and feature extraction, compared to the traditional text classification algorithm. Elghannam [34] constructs feature terms based on a novel bi-gram alphabet approach to construct feature terms.

## 3 Proposed Method

This paper proposes a method for creating a semantic representation that contains information about the relevance of a document to other documents in a topic using deep Siamese neural network. This method can be used to train a dense embedding for each document in the corpus. Documents' relevance to a common topic can be assessed using these embeddings. The deep Siamese neural network is composed of two multi-layer perceptron networks as sub-networks. The deepest layer of each MLP network is used as the document representation. Our network is trained to maximize the relevance score for a pair of documents with the same topic and minimize the relevance score for documents with different topics. In this section, we explain how conventional document representations are used as input for the Siamese neural network and the output layer of the network is used to calculate the relevance score for a pair of documents.

### 3.1 Improved document embedding

The inputs of the neural networks are the conventional document representations introduced in Section 2. These include TFIDF, LSA, LDA, and Glove word vector averages with different numbers of dimensions. In this section, we discuss the structure of the MLP sub-networks that are used to create document representations.

***Multi-layer perceptron***

We use two 1-layer perceptron networks as the sub-networks of the Siamese neural network. $W_1$ is the transformation function for building a deep representation for a document. In the application in this paper, both sub-networks are fed news documents as input. The same transformation function is used for both sub-networks. Therefore, each sub-network has the same number of layers and the same set of weights. These sub-networks extract abstract features from the input. Figure 1 shows the architecture of one of the MLP sub-networks.

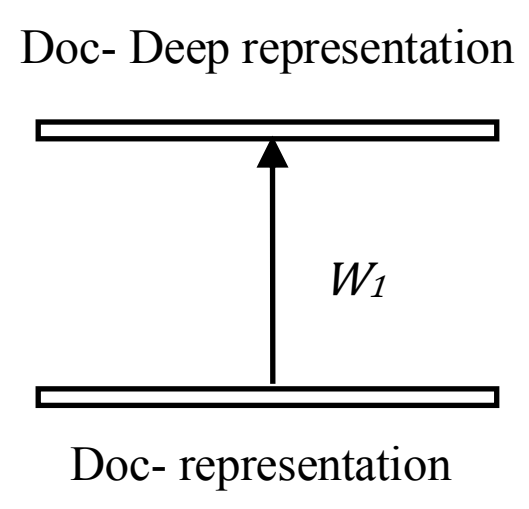

Figure 1: Schematic representation showing sub-network transformation function

### 3.2 Full Siamese Neural Network Architecture

In the Siamese neural network, the conventional document representation for two documents, *Doc$_a$ -representation* and *Doc$_b$ -representation*, are used as input to the MLP sub-networks. The sub-networks convert the document representations into two deep document representations, *$Doc_a$- Deep representation* and *$Doc_b$- Deep representation*, using the transformation function, $W_1$. *$Doc_a$- Deep representation* and *$Doc_b$- Deep representation* are combined using the combination function $W_{21}$. Then, the output layer $W_{22}$ is used to calculate the relevancy score for the two documents. Figure 2 depicts the architecture of our proposed network.

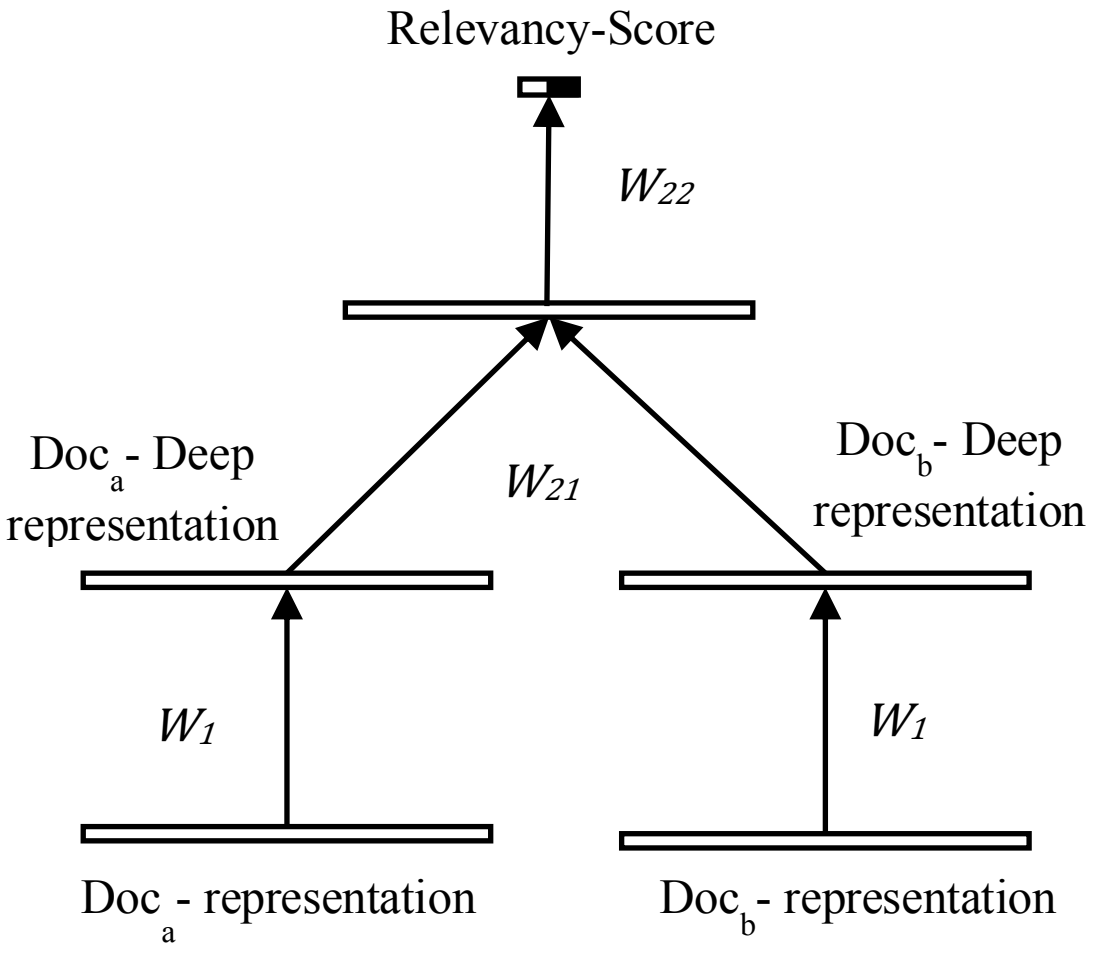

Figure 2: Schematic representation showing Siamese Neural Network

Mean square error (MSE) is used as the loss function for training the network. The training is stopped when MSE is minimized over the validation set. The parameters for training the network are the three sets of weights: $W_1$, $W_{21}$, $W_{22}$. Stochastic gradient descent is used as the optimization function during back-propagation.

The error for the input of each sub-network is different, so the gradient of the loss function with respect to $W_1$ for the input of each sub-network is calculated. The average of these values is used to update the sub-network weights. This is done so that both sub-networks have common weights.

The weight update for the transformation function, $W_1$, is calculated as:

$$\Delta_{W_1} = \frac{1}{2}\sum\left(\sum_{i=0}^{i=d_1} \delta_{W_{1a}}(i), \sum_{j=0}^{j=d_2} \delta_{W_{1b}}(j)\right) \tag{2}$$

where $\delta_{W_{1a}}(i)$ and $\delta_{W_{1b}}(j)$ are the deltas for layers of document *a* and document *b*. $d_1$ and $d_2$ are the number of layers of the two MLP sub-networks, which are both equal to 1 in this case.

We train different neural networks using each conventional representation as input and use the deepest layer of the MLP sub-networks as the improved document embeddings.

### 3.3 Text classification

In this section, we explain the employed classifiers. The classifiers used in this paper were: K-Nearest Neighbors (KNN), Support Vector Machine (SVM), Decision Tree (DT), Random Forest (RF) and Multi-Layer Perceptron (MLP).

The KNN algorithm finds the k-nearest neighbors to a test document among all the documents in the training set base on a similarity metric, here cosine similarity. After sorting the similarity values, a category is assigned to the test document based on the label of k-neighbors [35]. KNN classifier is expected to perform well on this task since the aim of the proposed method is to build representations that are close to each other with similar topics.

SVM creates a hyperplane that is designed to separate the data into two sets[36]. In a hard margin SVM, the hyperplane is designed to perfectly separate the training data so that all points on each side of the hyperplane belong to one class. In a soft margin SVM, the hyperplane is designed to minimize the number of points that are misclassified by the hyperplane. Non-linear separation can be achieved by mapping data to higher dimensions.

Decision tree classifiers create divisions of the feature space using the variables of the input data. Divisions are created by using a probabilistic loss function like Gini impurity or cross-entropy on the subset of data to test how well the division allows the model to separate the data. This creates a hierarchical decomposition of the data space [4]. Decision trees are prone to overfitting on data, so Random Forest has been proposed as an alternative. Random forest or random decision forest [37] is an ensemble learning method for classification. Random Forest uses bootstrap aggregating of Decision Trees. This reduces the variance of the classifier.

Multi-layer perceptron networks are a class of feed-forward deep neural networks with multiple hidden layers. In the training phase for this classifier the error is backpropagated and update the weights of the neural network[38].

## 4 Experiments

The text documents were preprocessed by eliminating non-alphabetic characters and converting the text to lowercase. The initial weights for the transformation function and combination function are set to random numbers between [-0.01, 0.01]. The activation function for the combination layer is Leaky ReLU (LReLU). Standard ReLU has a zero gradient for negative values, but LReLU has a non-zero gradient for negative values. This means that it is possible for the weights below the neuron with an LReLU activation function to be updated even when the input has a negative value. This function is shown in equation 3.

$$f(x) = max(\varepsilon x, x) \qquad (3)$$

The parameter ε is set to 0.1. The output layer activation function is Hyperbolic tangent. It is used to compute the relevancy score for the two input documents. For the $i^{th}$ iteration, the learning rate is given by equation 4.

$$lr_i = \frac{lr_0}{1 + \frac{i}{s}} \qquad (4)$$

where $lr_i$ is the learning rate of the $i^{th}$ iteration and $lr_0 = 0.0005$ is the initial learning rate. s = 100 is a hyper-parameter that sets the learning rate decay.

Dropout is a regularization technique that is used to prevent overfitting. It randomly eliminates some of the weights during the training process. In this paper %50 of the weights were eliminated during each training iteration. During the experiment a range of values for hyperparameters including learning rate and dropout were tested and the best test performance on the validation set is reported. Other deep neural networks including CNN[3], bidirectional_RNN[4], GRU[5], LSTM[6], RCNN[7] are also implemented and the performance are presented in the result section.

### 4.1 Dataset

The experiment was performed using data from the BBC news[8] corpus. We tested using all the data in the BBC news corpus. BBC news articles from 2004 to 2005 were gathered to create the BBC news corpus. The topics for the articles in the BBC news corpus are: business, entertainment, politics, sport, and tech.

---

[3] Convolutional Neural Network (CNN)
[4] Bidirectional Recurrent Neural Network (bidirectional_RNN)
[5] Gated Recurrent Unit (GRU)
[6] Long Short-Term Memory (LSTM)
[7] Recurrent Convolutional Neural Network (RCNN)
[8] http://mlg.ucd.ie/datasets/bbc.html

We split the data into train, test, and validation sets. Table 1 shows the number of samples in each sub-set for the BBC corpus.

Table 1: Corpus statistics

| | BBC |
|---|---|
| **Train** | 1803 |
| **Test** | 221 |
| **Validation** | 201 |

The Siamese neural network is designed to be trained for a binary task. The goal is to classify a pair of documents as relevant or non-relevant. Documents are considered to be relevant to each other if they have the same topic label [22]. Relevant samples are pairs of documents with the same topic. The relevancy score for relevant samples is equal to 1. For example: {(entertainment, entertainment), (trade , trade ), …} are relevant samples. Non-relevant pairs are pairs of documents with different topics. The relevancy score for relevant samples is equal to 0. For example: {(business, tech), (crude, money), …} are non-relevant samples. A large number of samples were generated by this approach to train and fine-tune deep neural networks.

More than 200,000 labeled document pairs were used to train the Siamese neural network to calculate the relevancy score for a pair of documents. Table 2 shows the number of document pair samples that were used to train the network.

Table 2: Pairs statistics

| #Samples | BBC |
|---|---|
| **Train** | 200000 |
| **Test** | 800 |
| **Validation** | 200 |

The training terminates when the error on the validation set starts to increase. The trained transformation function, $W_1$, can be used to create a dense semantic representation that embeds relevance of a document to a topic.

### 4.2 Results and Discussion

The goal of the experiment is to compare the semantic representations created using the deep Siamese neural network approach to the conventional representations discussed in section 2. The results of the text classification on different representations using the models discussed in section 3.3 are reported for comparison purposes.

Table 3 shows the performance of the text categorization by KNN with different number of neighbors on BBC dataset, respectively. We chose to report KNN separately as its classification method is aligned with our approach to building the representations. We aimed to build representations that are close to other representations for documents with the same topic in the vector space. Glove word vector average and the deep Glove word vector average outperform all other representations.

Table 3: F1 performance on BBC dataset - KNN classifier

| Dim | #neighbours | TF_IDF | LDA | LSA | AVG | Deep_TFIDF | Deep_LSA | Deep_AVG |
|---|---|---|---|---|---|---|---|---|
| *100* | 1 | 0.8105 | 0.4830 | 0.8059 | **0.9683** | 0.8817 | 0.8531 | 0.9680 |
| | 5 | 0.7991 | 0.3725 | 0.8032 | 0.9637 | 0.8400 | 0.8670 | **0.9726** |
| | 10 | 0.8080 | 0.5107 | 0.8083 | 0.9637 | 0.8449 | 0.8579 | **0.9726** |
| | 15 | 0.8139 | 0.4401 | 0.8233 | 0.9547 | 0.8444 | 0.8579 | **0.9726** |
| | 20 | 0.8153 | 0.6059 | 0.8237 | 0.9498 | 0.8492 | 0.8580 | **0.9683** |
| | | | | | | | | |
| *200* | 1 | 0.8814 | 0.4309 | 0.8814 | 0.9637 | 0.9503 | 0.7872 | **0.9772** |
| | 5 | 0.8773 | 0.4929 | 0.8773 | 0.9683 | 0.9364 | 0.7866 | **0.9772** |
| | 10 | 0.8909 | 0.3890 | 0.8909 | 0.9637 | 0.9318 | 0.7860 | **0.9818** |
| | 15 | 0.9089 | 0.2040 | 0.9042 | 0.9547 | 0.9271 | 0.7765 | **0.9728** |
| | 20 | 0.9085 | 0.3554 | 0.9085 | 0.9498 | 0.9271 | 0.7735 | **0.9728** |
| | | | | | | | | |
| *300* | 1 | 0.8909 | 0.4920 | 0.8956 | 0.9727 | 0.9230 | 0.7250 | **0.9773** |
| | 5 | 0.9135 | 0.4317 | 0.9135 | **0.9728** | 0.9180 | 0.7167 | 0.9683 |
| | 10 | 0.8999 | 0.4895 | 0.9044 | 0.9637 | 0.9270 | 0.7584 | **0.9728** |
| | 15 | 0.9043 | 0.4228 | 0.9043 | 0.9544 | 0.9361 | 0.7460 | **0.9728** |
| | 20 | 0.9088 | 0.2800 | 0.9088 | 0.9495 | 0.9269 | 0.7605 | **0.9728** |

For the BBC dataset, the best F1 score was %98.18. This was found using a deep Glove word vector average representation with 200 dimensions. This representation outperforms tf_idf by %8 and deep tf_idf by %3.

The conventional representations have better performance in higher dimensions. This is due to the exponential grow in number of variables to be optimized. Therefore, deep neural networks perform poorly.

Table 4 reports the text classification performance of the other four classifiers on different representations. Figure 3 shows the average of the performance over all the classifiers for different representations the BBC dataset. These experiments show that our deep representations outperform the conventional representations for most cases. For the BBC corpus, the results over all the deep representations are better than the conventional ones.

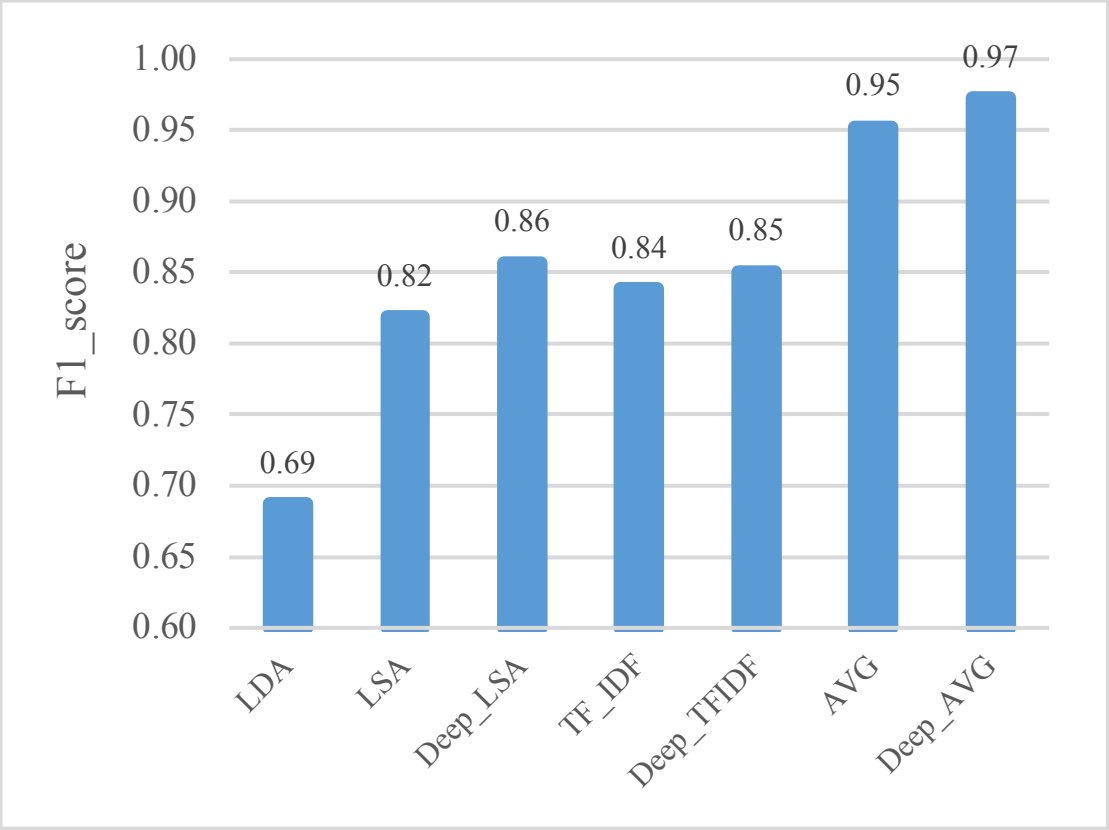

Figure 3: Average performance on different representations- BBC dataset

Table 4: F1 on BBC dataset classification

| Dim | Classifier | TF_IDF | LDA | LSA | AVG | Deep_TFIDF | Deep_LSA | Deep_AVG |
|---|---|---|---|---|---|---|---|---|
| *100* | SVM | 0.8441 | 0.7555 | 0.8535 | 0.9683 | 0.8719 | 0.8845 | **0.9728** |
| | DT | 0.7925 | 0.6527 | 0.7399 | 0.8963 | 0.8135 | 0.8082 | **0.9635** |
| | MLP | 0.8664 | 0.7783 | 0.8381 | 0.9637 | 0.8486 | 0.8574 | **0.9774** |
| | RF | 0.8807 | 0.7677 | 0.8488 | 0.9728 | 0.8670 | 0.8764 | **0.9774** |
| | | | | | | | | |
| *200* | SVM | 0.9271 | 0.7604 | 0.9271 | 0.9683 | 0.9321 | 0.8396 | **0.9818** |
| | DT | 0.7974 | 0.5964 | 0.8197 | 0.8682 | 0.8800 | 0.6144 | **0.9728** |
| | MLP | 0.9313 | 0.7756 | 0.9224 | 0.9590 | 0.9274 | 0.8113 | **0.9772** |
| | RF | 0.9272 | 0.7466 | 0.9275 | 0.9728 | 0.9226 | 0.7787 | **0.9818** |
| | | | | | | | | |
| *300* | SVM | 0.9405 | 0.8017 | 0.9360 | 0.9727 | 0.9087 | 0.8072 | **0.9772** |
| | DT | 0.8203 | 0.6122 | 0.8312 | 0.8587 | 0.8276 | 0.6932 | **0.9403** |
| | MLP | 0.9313 | 0.8584 | 0.9041 | 0.9729 | 0.9224 | 0.7925 | **0.9772** |
| | RF | 0.9358 | 0.7438 | 0.9368 | 0.9636 | 0.9220 | 0.7866 | **0.9728** |
| | | | | | | | | |
| *400* | SVM | **0.9543** | 0.7846 | 0.9543 | - | 0.9272 | 0.8521 | - |
| | DT | 0.7932 | 0.6056 | **0.8507** | - | 0.8043 | 0.6588 | - |
| | MLP | **0.9543** | 0.8712 | 0.9408 | - | 0.9412 | 0.8294 | - |
| | RF | **0.9498** | 0.7352 | 0.9241 | - | 0.9455 | 0.7937 | - |
| | | | | | | | | |
| *500* | SVM | **0.9500** | 0.7859 | 0.9500 | - | 0.9410 | 0.8350 | - |
| | DT | **0.8435** | 0.5941 | 0.8213 | - | 0.8316 | 0.6467 | - |
| | MLP | 0.9588 | 0.8125 | **0.9589** | - | 0.9454 | 0.8127 | - |
| | RF | **0.9498** | 0.6540 | 0.9285 | - | 0.9408 | 0.7708 | - |
| | | | | | | | | |
| *600* | SVM | 0.9501 | 0.8180 | 0.9501 | - | **0.9727** | 0.8494 | - |
| | DT | 0.8416 | 0.6782 | **0.8548** | - | 0.8329 | 0.6373 | - |
| | MLP | **0.9727** | 0.8825 | 0.9727 | - | 0.9682 | 0.8526 | - |
| | RF | 0.9498 | 0.7430 | 0.9371 | - | **0.9727** | 0.8259 | - |
| | | | | | | | | |
| *700* | SVM | **0.9501** | 0.8849 | **0.9501** | - | 0.9457 | 0.8488 | - |
| | DT | 0.8282 | 0.6344 | **0.8957** | - | 0.7635 | 0.6472 | - |
| | MLP | **0.9594** | 0.9412 | 0.9590 | - | 0.9498 | 0.8107 | - |
| | RF | **0.9634** | 0.7650 | 0.9551 | - | 0.9280 | 0.8119 | - |
| | | | | | | | | |
| *800* | SVM | 0.9411 | 0.8876 | 0.9411 | - | **0.9683** | 0.8174 | - |
| | DT | **0.8262** | 0.6408 | 0.8196 | - | 0.6891 | 0.6885 | - |
| | MLP | 0.9594 | 0.8957 | 0.9547 | - | **0.9729** | 0.8331 | - |
| | RF | 0.9452 | 0.7692 | 0.9372 | - | **0.9640** | 0.8095 | - |
| | | | | | | | | |
| *900* | SVM | **0.9545** | 0.8674 | 0.9545 | - | 0.9500 | 0.8452 | - |
| | DT | 0.7965 | 0.6167 | **0.8186** | - | 0.7477 | 0.6544 | - |
| | MLP | **0.9638** | 0.8990 | 0.9636 | - | 0.9544 | 0.8072 | - |
| | RF | **0.9591** | 0.7269 | 0.9505 | - | 0.9591 | 0.8125 | - |

The results of other deep neural networks using Glove word embedding with different dimensions as features are reported in Table 5. The best performance, %93.6, is reported by Recurrent-CNN.

Table 5: performance of deep neural network - BBC datasets

| Deep Neural Network | 100 | 200 | 300 |
|---|---|---|---|
| *CNN* | 0.9260 | 0.9204 | 0.9352 |
| *bidirectional_RNN* | 0.9357 | 0.9274 | 0.9183 |
| *GRU* | 0.9250 | 0.9223 | 0.8875 |
| *LSTM* | 0.9290 | 0.9212 | 0.9239 |
| *RCNN* | 0.9360 | 0.8880 | 0.9166 |

Table 6 shows the comparison of our approach to the other methods and state-of-the-art results.

Table 6: Comparing the performance of different representations on BBC dataset

| Method | F1_score | Method | F1_score |
|---|---|---|---|
| *NASARI+Babel2Vec* [32] | 0.9729 | *CNN* | 0.9352 |
| *Babel2Vec* [32] | 0.9765 | *bidirectional_RNN* | 0.9357 |
| *DDNN* [33] | 0.97 | *GRU* | 0.925 |
| *bi-gram alpabet* [34] | 0.926 | *LSTM* | 0.929 |
| *LDA* | 0.8584 | *RCNN* | 0.936 |
| *TF_IDF* | 0.9405 | ***Deep_TFIDF*** | 0.9503 |
| *LSA* | 0.9368 | ***Deep_LSA*** | 0.8845 |
| *AVG* | 0.9729 | ***Deep_AVG*** | **0.9818** |

The results show that our method outperform all the other approaches and achieve %98.18 of f1_score. Figure 4 and Table 5 show a T-SNE visualization for each representation for the test documents on the BBC dataset [39]. T-SNE works by fitting a probability distribution over data points in a high dimensional space and then mapping the points to a 2-D space in which the points are distributed such that they are plotted in a way that matches their distribution in higher dimensional space. This means that the plots can show non-linear clusters of points. Each topic label is represented by a different color. Each document is represented by a point in the T-SNE plot. The T-SNE plots for all the representations show some degree of separation between topics, but the T-SNE plot for the deep semantic representations using the Siamese neural network training approach shows that similar documents cluster better and are plotted very close to each other. Some of the results and figures are missing due to the high computational complexity.

## 5 Conclusions

In this paper, we propose a novel approach that learns a dense semantic representation that embeds relevancy information about the topic of documents. Using this approach, documents with the same topic have similar representations. These representations were created by training a Siamese neural network based on multi-layer perceptron sub-networks to classify document relevancy. The transformation function trained using this approach gives semantic representations for documents. In the experiment, we prove that incorporating deep neural networks improves performance in similarity search and document categorization. Using T-SNE plots, we also show that this improved representation results in better topic separation. Other tasks that use large document datasets could benefit from document representations created using deep Siamese neural networks. Performance could be improved by further optimization of hyper-parameters.

## References:

[1] D. Hoogeveen, L. Wang, T. Baldwin, and K. M. Verspoor, "Web Forum Retrieval and Text Analytics: A Survey," *Found. Trends® Inf. Retr.*, 2018.

[2] M. Jiang *et al.*, "Text classification based on deep belief network and softmax regression," *Neural Comput. Appl.*, 2018.

[3] S. Lai, L. Xu, K. Liu, and J. Zhao, "Recurrent convolutional neural networks for text classification," in *Proceedings of the 29th AAAI Conference on Artificial Intelligence (AAAI-2015)*, 2015.

[4] C. Aggarwal and C. Zhai, "A Survey of Text Classification Algorithms," *Min. Text Data*, pp. 163–222, 2012.

[5] K. Kowsari, D. E. Brown, M. Heidarysafa, K. Jafari Meimandi, M. S. Gerber, and L. E. Barnes, "HDLTex: Hierarchical Deep Learning for Text Classification," in *Proceedings - 16th IEEE International Conference on Machine Learning and Applications, ICMLA 2017*, 2018.

[6] A. M. Rinaldi, "A content-based approach for document representation and retrieval," 2008.

[7] R. Williams, "A computational effective document semantic representation," in *Proceedings of the 2007 Inaugural IEEE-IES Digital EcoSystems and Technologies Conference, DEST 2007*, 2007.

[8] K. B. Shaban, "A semantic approach for document clustering," *J. Softw.*, 2009.

[9] R. Socher, C. D. C. Manning, and A. Y. A. Ng, "Learning continuous phrase representations and syntactic parsing with recursive neural networks," *Proc. NIPS-2010 Deep Learn. Unsupervised Featur. Learn. Work.*, pp. 1–9, 2010.

[10] K. S. Tai, R. Socher, and C. D. Manning, "Improved Semantic Representations From Tree-Structured Long Short-Term Memory Networks," *Proc. ACL*, pp. 1556–1566, 2015.

[11] Q. Le and T. Mikolov, "word2vec-v3," *Int. Conf. Mach. Learn. - ICML 2014*, vol. 32, pp. 1188–1196, 2014.

[12] J. BROMLEY *et al.*, "SIGNATURE VERIFICATION USING A 'SIAMESE' TIME DELAY NEURAL NETWORK," *Int. J. Pattern Recognit. Artif. Intell.*, vol. 07, no. 04, pp. 669–688, 1993.

[13] E. Gharavi, R. Silwal, M. S. Gerber, and H. Veisi, "Siamese Discourse Structure Recursive Neural Network for Semantic Representation," in *Proceedings - 13th IEEE International Conference on Semantic Computing, ICSC 2019*, 2019.

[14] G. Salton and C. Buckley, "Term-weighting approaches in automatic text retrieval," *Inf. Process. Manag.*, vol. 24, no. 5, pp. 513–523, 1988.

[15] S. T. Dumais, G. W. Furnas, T. K. Landauer, S. Deerwester, and R. Harshman, "Using latent semantic analysis to improve access to textual information," in *Proceedings of the SIGCHI conference on Human factors in computing systems - CHI '88*, 1988, pp. 281–285.

[16] D. M. Blei, A. Y. Ng, and M. I. Jordan, "Latent Dirichlet Allocation David," *J. Mach. Learn. Res.*, vol. 3, no. Jan, pp. 993–1022, 2003.

[17] E. Gharavi, K. Bijari, K. Zahirnia, and H. Veisi, "A deep learning approach to Persian plagiarism detection," in *CEUR Workshop Proceedings*, 2016, vol. 1737, pp. 154–159.

[18] E. Gharavi, H. Veisi, K. Bijari, and K. Zahirnia, "A Fast Multi-level Plagiarism Detection Method Based on Document Embedding Representation," in *Lecture Notes in Computer Science (including subseries Lecture Notes in Artificial Intelligence and Lecture Notes in Bioinformatics)*, 2018.

[19] E. Gharavi, H. Veisi, and P. Rosso, "Scalable and language-independent embedding-based approach for plagiarism detection considering obfuscation type: no training phase," *Neural Comput. Appl.*, 2019.

[20] J. Pennington, R. Socher, and C. D. Manning, "GloVe: Global Vectors for Word Representation," *Proc. 2014 Conf. Empir. Methods Nat. Lang. Process.*, pp. 1532–1543, 2014.

[21] R. Salakhutdinov and G. Hinton, "Semantic hashing," *Int. J. Approx. Reason.*, vol. 50, no. 7, pp. 969–978, 2009.

[22] Q. Wang, D. Zhang, and L. Si, "Semantic hashing using tags and topic modeling," in *Proceedings of the 36th international ACM SIGIR conference on Research and development in information retrieval - SIGIR '13*, 2013, p. 213.

[23] M. A. Livermore, F. Dadgostari, M. Guim, P. Beling, and D. Rockmore, "Law Search as Prediction," *Virginia Public Law Leg. Theory Res. Pap.*, no. 2018–61, 2018.

[24] P. Huang *et al.*, "Learning Deep Structured Semantic Models for Web Search using Clickthrough Data," *22nd ACM Int. Conf. Conf. Inf. Knowl. Manag.*, pp. 2333–2338, 2013.

[25] C. Lioma, B. Larsen, and W. Lu, "Rhetorical Relations for Information Retrieval," in *Proceedings of the 35th International ACM SIGIR Conference on Research and Development in Information Retrieval*,

2012, pp. 931–940.

[26] J. Mueller, "Siamese Recurrent Architectures for Learning Sentence Similarity," *Proc. 30th Conf. Artif. Intell. (AAAI 2016)*, no. 2012, pp. 2786–2792, 2016.

[27] Y. Ji and N. Smith, "Imported from Neural Discourse Structure for Text Categorization. (arXiv:1702.01829v1 [cs.CL]) http://arxiv.org/abs/1702.01829," *Preprint*, 2017.

[28] W. Yin, H. Schütze, B. Xiang, and B. Zhou, "Abcnn: Attention-based convolutional neural network for modeling sentence pairs," *arXiv Prepr. arXiv1512.05193*, 2015.

[29] and A. I. Zhiguo Wang, Haitao Mi, "Semi-supervised clustering for short text via deep representation learning," in *The 20th SIGNLL Conference on Computational Natural Language Learning (CoNLL)*, 2016.

[30] A. Wang, Zhiguo and Mi, Haitao and Ittycheriah, "Semi-supervised clustering for short text via deep representation learning," *arXiv Prepr. arXiv1602.06797*, 2016.

[31] S. R. Bowman, L. Vilnis, O. Vinyals, A. M. Dai, R. Jozefowicz, and S. Bengio, "Generating sentences from a continuous space," *arXiv Prepr. arXiv1511.06349*, 2015.

[32] R. A. Sinoara, J. Camacho-Collados, R. G. Rossi, R. Navigli, and S. O. Rezende, "Knowledge-enhanced document embeddings for text classification," *Knowledge-Based Syst.*, 2019.

[33] W. Aziguli *et al.*, "A Robust Text Classifier Based on Denoising Deep Neural Network in the Analysis of Big Data," *Sci. Program.*, 2017.

[34] F. Elghannam, "Text representation and classification based on bi-gram alphabet," *Journal of King Saud University - Computer and Information Sciences*, 2019.

[35] S. Jiang, G. Pang, M. Wu, and L. Kuang, "An improved K-nearest-neighbor algorithm for text categorization," *Expert Syst. Appl.*, 2012.

[36] H. Bo, G., & Xianwu, "SVM Multi-Class Classification [J]," *J. Data Acquis. Process.*, 2006.

[37] T. K. Ho, "Random decision forests," in *Proceedings of the International Conference on Document Analysis and Recognition, ICDAR*, 1995.

[38] D. E. Rumelhart, G. E. Hinton, and R. J. Williams, "Learning Internal Representations by Error Propagation," in *Readings in Cognitive Science: A Perspective from Psychology and Artificial Intelligence*, 2013.

[39] G. H. Laurens van der Maaten, "Visualizing Data using t-SNE," *J. Mach. Learn. Res.*, vol. 9, no. 85, pp. 2579–2605, 2008.

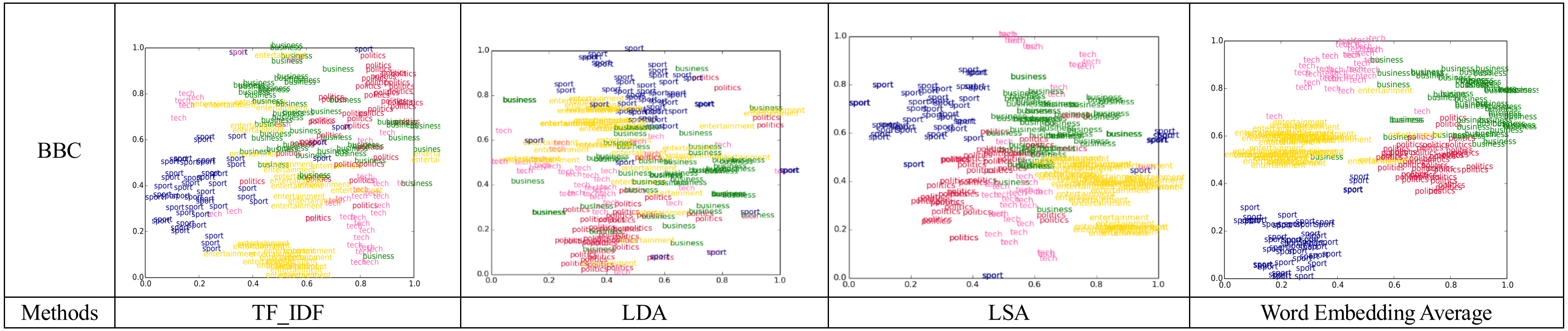

Figure 4: A 2-dimensional embedding for representations of test documents by different shallow methods using t-sne. See in color for better visualization.

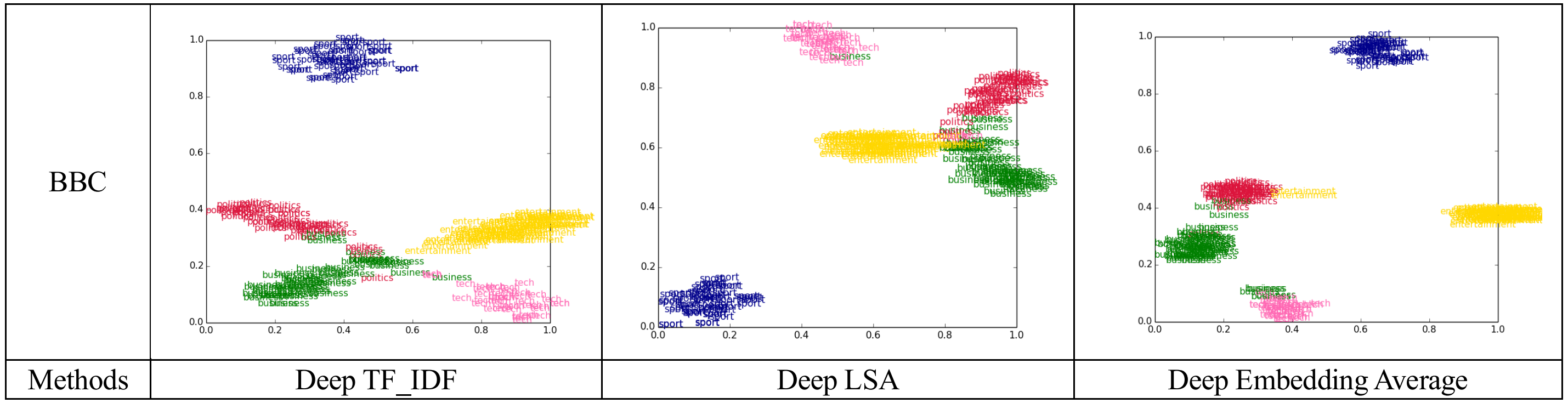

Figure 5: A 2-dimensional embedding for representations of test documents by different deep methods using t-sne. See in color for better visualization.